# Optimization With Parity Constraints: From Binary Codes to Discrete Integration*

**Stefano Ermon, Carla P. Gomes**
Department of Computer Science
Cornell University, Ithaca, NY, USA
{ermonste,gomes}@cs.cornell.edu

**Ashish Sabharwal**
IBM Watson Research Center
Yorktown Heights, NY, USA
ashish.sabharwal@us.ibm.com

**Bart Selman**
Dept. of Computer Science
Cornell University, Ithaca, USA
selman@cs.cornell.edu

## Abstract

Many probabilistic inference tasks involve summations over exponentially large sets. Recently, it has been shown that these problems can be reduced to solving a polynomial number of MAP inference queries for a model augmented with randomly generated parity constraints. By exploiting a connection with max-likelihood decoding of binary codes, we show that these optimizations are computationally hard. Inspired by iterative message passing decoding algorithms, we propose an Integer Linear Programming (ILP) formulation for the problem, enhanced with new sparsification techniques to improve decoding performance. By solving the ILP through a sequence of LP relaxations, we get both lower and upper bounds on the partition function, which hold with high probability and are much tighter than those obtained with variational methods.

## 1 INTRODUCTION

Discrete probabilistic graphical models [18, 31] are often defined up to a normalization factor involving a summation over an exponentially large combinatorial space. Computing these factors is an important problem, as they are needed, for instance, to evaluate the probability of evidence, rank two alternative models, and learn parameters from data. Unfortunately, computing these discrete integrals exactly in very high dimensional spaces quickly becomes intractable, and approximation techniques are often needed. Among them, sampling and variational methods are the most popular approaches. Variational inference problems are typically solved using message passing techniques, which are often guaranteed to converge to some local minimum [30, 31], but without guarantees on the quality of the solution found. Markov Chain Monte Carlo [17, 21, 32] and Importance Sampling techniques [10, 11, 13] are asymptotically correct, but the number of samples required to obtain a statistically reliable estimate can grow exponentially in the worst case.

Recently, Ermon et al. [6] introduced a new technique called WISH which comes with provable (probabilistic) guarantees on the approximation error. Their method combines combinatorial optimization techniques with the use of universal hash functions to uniformly partition a large combinatorial space, originally introduced by Valiant and Vazirani to study the Unique Satisfiability problem and later exploited by Gomes et al. [13, 14] for solution counting. Specifically, they show that one can obtain the intractable normalization constant (partition function) of a graphical model within any desired degree of accuracy, by solving a polynomial number of MAP queries for the original graphical model augmented with randomly generated parity constraints as evidence. Although MAP inference is NP-hard and thus also intractable, this is a significant step forward as counting problems such as estimating the partition function are #-P hard, a complexity class believed to be significantly harder than NP.

In this work, we investigate the class of MAP inference queries with random parity constraints arising from the WISH scheme. These optimization problems turn out to be intimately connected with the fundamental problem of maximum likelihood decoding of a binary code [3, 29]. We leverage this connection to show that the inference queries generated by WISH are NP-hard to solve and to approximate, even for very simple graphical models. Although generally hard in the worst case, message passing and related linear programming techniques [7] are known to be very successful in practice in decoding certain types of codes such as low density parity check (LDPC) codes [8].

*This work was supported by NSF Grant 0832782.

Inspired by the success of these methods, we formulate the MAP inference queries generated by WISH as Integer Linear Programs (ILP). Unfortunately, such queries are typically harder than traditional decoding problems because they involve more complex probabilistic models, and because universal hash functions naturally give rise to very "dense" parity constraints. To address this issue, we propose a technique to construct equivalent but sparser (and empirically easier to solve) parity constraints. Further, we introduce a more general version of WISH that relies *directly* on arbitrarily sparse parity constraints, thus giving rise to easier to solve MAP queries but providing weaker, one-sided guarantees on the approximation error for the partition function.

Our ILP formulation with sparsification techniques provides very good lower bounds on the partition function, while at the same time providing also upper bounds based on solving a sequence of LP relaxations. These upper bounds are much tighter than those obtained by tree decomposition and convexity [30]. This is a significant advance, because other state-of-the-art sampling based algorithms [10, 11, 13, 32] can usually provide probabilistic guarantees on lower bounds, but are not able to reason at all about upper bounds.

## 2 PROBLEM STATEMENT

We consider a discrete probabilistic graphical model [31] with $n = |V|$ random variables $\{x_i, i \in V\}$ where each random variable $x_i$ takes values in a finite set $\mathcal{X}_i$. We consider a factor graph representation for a joint probability distribution over elements $x \in \mathcal{X} = \mathcal{X}_1 \times \cdots \times \mathcal{X}_n$ (also referred to as **configurations**)

$$p(x) = \frac{1}{Z} \prod_{\alpha \in \mathcal{I}} \psi_\alpha(\{x\}_\alpha) \quad (1)$$

This is a compact representation for $p(x)$, which is defined as the product of non-negative factors $\psi_\alpha : \{x\}_\alpha \mapsto \mathbb{R}^+$, where $\mathcal{I}$ is an index set and $\{x\}_\alpha \subseteq V$ a subset of variables the factor $\psi_\alpha$ depends on. $Z$ is a normalization constant known as *partition function* ensuring the probabilities sum up to one. Formally the partition function $Z$ is defined as

$$Z = \sum_{x \in \mathcal{X}} \prod_{\alpha \in \mathcal{I}} \psi_\alpha(\{x\}_\alpha) = \sum_{x \in \mathcal{X}} w(x) \quad (2)$$

where for compactness we have introduced a weight function $w : \mathcal{X} \to \mathbb{R}^+$ that assigns to each configuration $x \in \mathcal{X}$ its unnormalized probability, namely

$$w(x) = \prod_{\alpha \in \mathcal{I}} \psi_\alpha(\{x\}_\alpha) \quad (3)$$

Computing the partition function $Z$ is a #-P complete, intractable problem because it involves a sum over an exponentially large number of configurations. However, the partition function is a key property of a graphical model, needed e.g. to actually evaluate the probability of a configuration $x$ under $p$. In this paper, we will focus on approximate techniques to estimate and bound this quantity. For simplicity, we consider the case of binary variables where $x_i \in \mathcal{X}_i = \{0, 1\}$. The general case can be encoded using a bit representation and binary variables.

## 3 BACKGROUND

This paper extends previous work by Ermon et al. [6] who introduced an algorithm called WISH to estimate the partition function (2). WISH is a randomized approximation algorithm that gives a constant factor approximation of $Z$ with high probability. It involves solving a polynomial number of MAP inference queries for the graphical model conditioned on randomly generated evidence based on universal hashing.

### 3.1 FAMILIES OF HASH FUNCTIONS

A key ingredient of the WISH algorithm is the concept of **pairwise independent** hashing, originally introduced by Carter and Wegman [5] and later recognized as a tool that "should belong to the bag of tricks of every computer scientist" [33]. There are several in-depth expositions of the topic [cf. 12, 27, 28]. Here we will also make use of a weaker notion of hashing, called **uniform** hashing and defined as follows:

**Definition 1.** A family of functions $\mathcal{H} = \{h : \{0,1\}^n \to \{0,1\}^m\}$ is called **uniform** if for $H \in_R \mathcal{H}$ it holds that $\forall x \in \{0,1\}^n$, the random variable $H(x)$ is uniformly distributed in $\{0,1\}^m$.

Here we use the notation $H \in_R \mathcal{H}$ to denote $H$ being chosen uniformly at random from $\mathcal{H}$.

**Definition 2.** A family of functions $\mathcal{H} = \{h : \{0,1\}^n \to \{0,1\}^m\}$ is called **pairwise independent** if it is uniform and for $H \in_R \mathcal{H}$ it holds that $\forall x_1, x_2 \in \{0,1\}^n$ with $x_1 \neq x_2$, the random variables $H(x_1)$ and $H(x_2)$ are independent.

Many constructions of pairwise independent hash functions are known. A simple and well-known one was used by Ermon et al.:

**Proposition 1** ([6]). *Let $A \in \{0,1\}^{m \times n}$, $b \in \{0,1\}^m$. The family $\mathcal{H}^{n,m} = \{h_{A,b}(x) : \{0,1\}^n \to \{0,1\}^m\}$ where $h_{A,b}(x) = Ax + b \mod 2$ is a family of pairwise independent hash functions.*

**Algorithm 1** WISH $(w, n = \log_2 |\mathcal{X}|, \delta, \alpha, \{\mathcal{H}^{n,i}\})$

$T \leftarrow \left\lceil \frac{\ln(1/\delta)}{\alpha} \ln n \right\rceil$
**for** $i = 0, \cdots, n$ **do**
    **for** $t = 1, \cdots, T$ **do**
        Sample hash function $h_{A,b}^i$ uniformly from $\mathcal{H}^{n,i}$
        $w_i^t \leftarrow \max_\sigma w(\sigma)$ subject to $h_{A,b}^i(\sigma) = \mathbf{0}$
    **end for**
    $M_i \leftarrow \text{Median}(w_i^1, \cdots, w_i^T)$
**end for**
Return $M_0 + \sum_{i=0}^{n-1} M_{i+1} 2^i$

## 3.2 THE WISH ALGORITHM FOR DISCRETE INTEGRATION

The basic idea behind WISH is to (implicitly) randomly partition the space of all possible configurations by universally hashing configurations into $2^m$ buckets. This step is achieved using randomly generated *parity constraints* of the form $Ax = b \mod 2$, which may also be viewed as logical XOR operations acting on the binary variables of the problem: $A_{i1}x_1 \oplus A_{i2}x_2 \oplus \cdots \oplus A_{in}x_n = b_i$. A combinatorial optimization solver is then used to find a configuration with the largest weight within a *single* bucket. This corresponds to solving a MAP query, i.e., solving an optimization problem subject to (randomly generated) parity constraints. By varying the number of buckets and repeating the process a small number of times, this strategy provably yields an estimate of the intractable normalization factor (2) within any desired degree of accuracy, with high probability and using only a polynomial number of MAP queries. For completeness, we provide the pseudocode for WISH as Algorithm 1, modified to have the hash families $\mathcal{H}^{n,i}, i \in \{0, 1, \ldots, n\}$, as parameters whose variations we will consider later[1]. We will write WISH($\{\mathcal{H}^{n,i}\}$) when the values of the other parameters are implicit.

Although MAP inference itself is an NP-hard problem, this strategy is still desirable considering that computing $Z$ is a #P-hard problem, a complexity class believed to be even harder than NP. In practice, Ermon et al. [6] showed that the resulting MAP inference can be solved reasonably well using a state-of-the-art MAP inference engine called Toulbar [1], which was extended with custom propagators for parity constraints.

**Theorem 1** ([6]). *For any $\delta > 0$, positive constant $\alpha \leq 0.0042$, and the hash families $\mathcal{H}^{n,i}$ given by Proposition 1, WISH($\{\mathcal{H}^{n,i}\}$) makes $\Theta(n \ln n \ln 1/\delta)$ MAP queries and, with probability at least $(1 - \delta)$, outputs a 16-approximation of $Z = \sum_{\sigma \in \mathcal{X}} w(\sigma)$.*

---
[1] For $i = 0$, $h_{A,b}^i \equiv \mathbf{0}$ and no constraint is added.

Further, even if the MAP instances in the inner loop of Algorithm 1 are not solved to optimality, the output of the algorithm using suboptimal MAP solutions is an *approximate lower bound* for $Z$ (specifically, no more than $16Z$) with probability at least $(1 - \delta)$. If suboptimal solutions are within a constant factor $L$ of the optimal, then the output is a $16L$-approximation of $Z$ with probability at least $(1-\delta)$ [6]. Similarly, if one has access to upper bounds to the values of the MAP instances, the output of the algorithm using these upper bounds is an *approximate upper bound* (specifically, at least $1/16Z$) for $Z$ with probability at least $(1 - \delta)$.

## 3.3 IMPROVING WISH: HASHING USING TOEPLITZ MATRIX

The performance of Algorithm 1 can be improved by constructing pairwise independent hash functions not by choosing $A \in_R \{0,1\}^{i \times n}$ but rather letting $A$ be a random $i \times n$ Toeplitz matrix [24]. Specifically, the first column and row of $A$ are filled with uniform i.i.d. Bernoulli variables in $\{0,1\}$. The value of each entry is then copied into the corresponding descending top-left to bottom-right diagonal. This process requires $n + i - 1$ random bits rather than $ni = O(n^2)$. Let $\mathcal{T}(m,n) \subseteq \{0,1\}^{m \times n}$ be the set of $m \times n$ Toeplitz matrices with $0, 1$ entries. Then:

**Proposition 2** ([12, 27]). *Let $A \in \mathcal{T}(m,n)$, $b \in \{0,1\}^m$. The family $\mathcal{H}_{\mathcal{T}}^{n,m} = \{h_{A,b}(x) : \{0,1\}^n \to \{0,1\}^m\}$ where $h_{A,b}(x) = Ax + b \mod 2$ is a family of pairwise independent hash functions.*

WISH($\{\mathcal{H}_{\mathcal{T}}^{n,m}\}$) still provides the same theoretical guarantees as Theorem 1 but has a more deterministic and stable behavior as it requires only $\Theta(n^2 \log n)$ random bits rather than $\Theta(n^3 \log n)$.

## 4 CONNECTIONS WITH CODING THEORY

For a problem with $n$ binary variables, WISH requires solving $\Theta(n \log n)$ optimization instances. If these optimizations could be approximated (within a constant factor of the true optimal value) in polynomial time, this would give rise to a polynomial time algorithm that gives, with high probability, a constant factor approximation for the original counting problem. Note that this is a reasonable assumption, because perhaps the most interesting #-P complete counting problems are those whose corresponding decision problem are easy, e.g. counting weighted matchings in a graph (computing the permanent). A natural question arises: *are there interesting counting problems for which we can approximate $\max_\sigma w(\sigma)$ subject to $A\sigma = b \mod 2$ in polynomial time?*

To shed some light on this question, we show a connection with a decision problem arising in coding theory:

**Definition 3** (MAXIMUM-LIKELIHOOD DECODING). *Given a binary $m \times n$ matrix $A$, a vector $b \in \{0,1\}^m$, and an integer $w > 0$, is there a vector $z \in \{0,1\}^n$ of Hamming weight $\leq w$, such that $Az = b \mod 2$?*

As noted by Vardy [29], Berlekamp et al. [3] showed that this problem is NP-complete with a reduction from 3-DIMENSIONAL MATCHING. Further, Stern [26] and Arora et al. [2] proved that even approximating within any constant factor the solution to this problem is NP-hard.

These hardness results restrict the kind of problems we can hope to solve in our setting, which is more general. In fact, we can define a graphical model with single variable factors $\psi_i(x_i) = \exp(-x_i)$ for $x_i \in \{0,1\}$. Let $\mathcal{S} = \{x \in \{0,1\}^n : Ax = b \mod 2\}$. Then

$$\max_{x \in \mathcal{S}} w(x) = \max_{x \in \mathcal{S}} \prod_{i=1}^n \psi_i(x_i) = \exp\left(\max_{x \in \mathcal{S}} \sum_{i=1}^n \log \psi_i(x_i)\right)$$
$$= \exp\left(\max_{x \in \mathcal{S}} -H(x)\right) = \exp\left(-\min_{x \in \mathcal{S}} H(x)\right)$$

where $H(x)$ is the Hamming weight of $x$. Thus, MAXIMUM-LIKELIHOOD DECODING of a binary code is a special case of MAP inference subject to parity constraints, but on a simple (disconnected) factor graph with factors acting only on single variable nodes. Intuitively, in the context of coding theory, there is a variable for each transmitted bit, and factors capture the probability of a transmission error on each bit. Thus there are no interactions between the variables, except for the ones introduced by the parity constraints $Ax = b \mod 2$, while in our context we allow for more complex probabilistic dependencies between variables specified as in Eq. (1). We therefore have the following theorem:

**Theorem 2.** *Given a binary $m \times n$ matrix $A$, a vector $b \in \{0,1\}^m$, and $w(x)$ as in Equation (3), the following optimization problem*

$$\max_{x \in \{0,1\}^n} \log w(x) \text{ subject to } Ax = b \mod 2$$

*is NP-hard to solve and to approximate within any constant factor.*

Connections with coding theory is even deeper, and is not just an artifact of the particular hash function construction used. In fact, there is an intimate connection and a correspondence between universal hash functions and (binary) codes, where one can construct hash functions from binary codes and vice versa [27].

## 4.1 MESSAGE PASSING DECODING

Iterative Message Passing (MP) methods are among the most widely used decoding techniques. Although the decoding problem is computationally intractable, they usually have very good performance in practice [7, 19]. Since we can represent parity constraints as additional factors in our original factor graph model, MP techniques can also be heuristically applied to solve the more general MAP inference queries with parity constraints generated by WISH. Specifically, although a parity constraint over $k$ variables would require a conditional probability table (CPT) of size $2^k$ to be specified, efficient Dynamic-Programming-based updates for parity constraints are known, see e.g. [19]. These updates have complexity which is linear in $k$, and thus, by representing parity constraints implicitly, we can directly use these techniques.

## 5 INTEGER PROGRAMMING FORMULATION

The NP-hard combinatorial optimization problem $\max_\sigma w(\sigma)$ subject to $A\sigma = b \mod 2$ can be formulated as an Integer Program [4]. This is a promising approach because Integer Linear Programs and related Linear programming (LP) relaxations have been shown to be a very effective at decoding binary codes by Feldman et al. [7]. Further, the empirically successful iterative message-passing decoding algorithms are closely related to LP relaxations of certain Integer Programs, either because they are directly trying to solve an LP or its dual like the MPLP and TRWBP [9, 25, 30], or attempting to approximately solve a variational problem over the same polytope like Loopy Belief Propagation [31].

### 5.1 MAP INFERENCE AS AN ILP

For simplicity, we consider the case of binary factors (pairwise interactions between variables), where equation (3) simplifies to $w(x) = \prod_{i \in V} \psi_i(x_i) \prod_{(i,j) \in E} \psi_{ij}(x_i, x_j)$ for some edge set $E$. Rewriting in terms of the logarithms, the unconstrained MAP inference problem can be stated as $\max_{x \in \{0,1\}^n} \sum_{i \in V} \theta_i(x_i) + \sum_{(i,j) \in E} \theta_{ij}(x_i, x_j)$ which can be written as an Integer Linear Program using binary indicator variables $\{\mu_i, i \in V\}$ and $\{\mu_{ij}(x_i, x_j), (i,j) \in E, x_i \in \{0,1\}, x_j \in \{0,1\}\}$ as follows [31]:

$$\max_{\mu_i, \mu_{ij}(x_i, x_j)} \sum_{i \in V} \theta_i(1)\mu_i + \theta_i(0)(1-\mu_i) +$$
$$\sum_{(i,j) \in E} \sum_{x_i, x_j} \theta_{ij}(x_i, x_j)\mu_{i,j}(x_i, x_j)$$

subject to

$$\forall i \in V, (i,j) \in E, \quad \sum_{x_j \in \{0,1\}} \mu_{i,j}(0, x_j) = 1 - \mu_i$$

$$\forall i \in V, (i,j) \in E, \quad \sum_{x_j \in \{0,1\}} \mu_{i,j}(1, x_j) = \mu_i$$

$$\forall i \in V, (i,j) \in E, \quad \sum_{x_i \in \{0,1\}} \mu_{i,j}(x_i, 0) = 1 - \mu_j$$

$$\forall i \in V, (i,j) \in E, \quad \sum_{x_i \in \{0,1\}} \mu_{i,j}(x_i, 1) = \mu_j$$

## 5.2 PARITY CONSTRAINTS

There are several possible encodings for the parity constraints $A\sigma = b \mod 2$, defining the so called **parity polytope** over $\sigma \in \mathbb{R}^n$. We summarize them next. Let $\mathcal{J}$ be the set of parity constraints (one entry per row of $A$). Let $\mathcal{N}(j)$ be the set of variables the $j$-th parity constraint depends on, namely the indexes of the non-zero columns of the $j$-th row of $A$[2]. We'll refer to $|\mathcal{N}(j)|$ as the length of the $j$-th XOR.

### 5.2.1 Exponential polytope representation

The simplest encoding is due to Jeroslow [16]. It requires that for all $j \in \mathcal{J}$, $S \subseteq \mathcal{N}(j)$, and $|S|$ odd, the following should hold

$$\sum_{i \in S} \mu_i + \sum_{i \in (\mathcal{N}(j) \setminus S)} (1 - \mu_i) \leq |\mathcal{N}(j)| - 1$$

Clearly, this requires a number of constraints that is exponential in the length of the XOR.

Another representation exponential in the length of the parity constraint is due to Feldman et al. [7]. For each $S$ in the set $E_j = \{S \subseteq \mathcal{N}(j) : |S| \text{ even}\}$ there is an extra binary variable $w_{j,S} \in \{0,1\}$. It requires $\forall j \in \mathcal{J}, \sum_{S \in E_j} w_{j,S} = 1$ and $\forall j \in \mathcal{J}, \forall i \in \mathcal{N}(j), \mu_i = \sum_{S \in E_j : i \in S} w_{j,S}$.

### 5.2.2 Compact polytope representation

Yannakakis [34] introduced the following compact representation which requires only $O(n^3)$ variables and constraints, where $n$ is the number of variables. For each constraint $j$, define $T_j = \{0, 2, \cdots, 2\lfloor|\mathcal{N}(j)|/2\rfloor\}$ as the set of even numbers between 0 and $|\mathcal{N}(j)|$.

- for all $j \in \mathcal{J}$ and for all $k \in T_j$ we have a binary variable $\alpha_{j,k} \in \{0,1\}$

- for all $j \in \mathcal{J}$ and for all $k \in T_j$ and for all $i \in \mathcal{N}(j)$ we have a binary variable $z_{i,j,k} \in \{0,1\}$, $0 \leq z_{i,j,k} \leq \alpha_{j,k}$

---
[2]To represent the desired parity of the $j$-th constraint imposed by $b_j$ we use a dummy variable $d = 1$, and include $d$ in $\mathcal{N}(j)$ whenever $b_j = 1$.

Then the following constraints are enforced:

$$\forall i \in V, j \in \mathcal{N}(i), \quad \mu_i = \sum_{k \in T_j} z_{i,j,k}$$

$$\forall j \in \mathcal{J}, \quad \sum_{k \in T_j} \alpha_{j,k} = 1$$

$$\forall j \in \mathcal{J}, \forall k \in T_j, \quad \sum_{i \in \mathcal{N}(j)} z_{i,j,k} = k\alpha_{j,k}$$

For any set of parity constraints, these 3 encodings are equivalent, in the sense that the subset of $\{\mu_i\}$ satisfying the constraints is the same [7]. Thus, the corresponding MAP inference problems are also equivalent, as the objective function, by expressing each $\mu_{i,j}$ in terms of the $\{\mu_i\}$ variables, can be re-written as a (possibly non-linear) function of only $\{\mu_i\}$.

## 5.3 SOLVING INTEGER PROGRAMS

Solving ILPs typically relies on solving a sequence of Linear Programming (LP) relaxations obtained by relaxing the integrality constraints. The solution to the relaxation provides an upper bound to the original integer maximization problem. Since LP can be solved in polynomial time, using Theorem 1 and following remarks we have a **polynomial time** method to obtain approximate upper bounds on the partition function which hold with high probability, although without tightness guarantees. Notice that upper bounds of this form could also be obtained using message passing techniques such as MPLP or TRWBP [9, 25, 30], which can also provide upper bounds to the values of the MAP inference queries in the inner loop of WISH.

IP solvers such as IBM ILOG CPLEX Optimization Studio [15] solve a sequence of LP relaxations based on branching on the problems's variables, iteratively improving the upper bound and keeping track of the best integer solution found, until lower and upper bounds match. Thus, one advantage of using an IP solver over standard Message Passing techniques is that the upper and lower bounds improve over time, and it is guaranteed to eventually provide an optimal solution for the original integer problem. In Figure 1 we plot the upper bound reported by CPLEX as a function of runtime for a random $10 \times 10$ Ising model with mixed interactions. It's clear that there is quickly a dramatic improvement over the value of the basic LP relaxation, which is the value reported by CPLEX around time zero, and that the upper bound keeps improving although at a slower rate. We note that other techniques such as by Sontag et al. [25] could also be used to iteratively tighten the LP relaxation, and might lead to better scaling behavior on certain classes of very large problems [35].

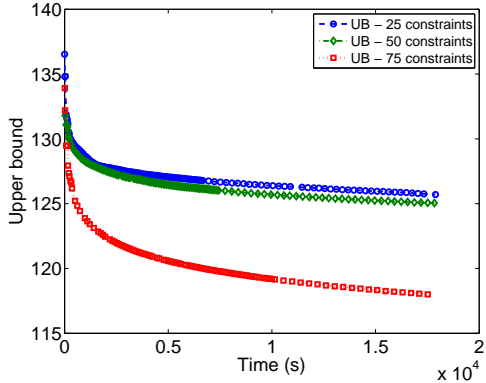

Figure 1: Upper bound as a function of runtime.

## 5.4 INDUCING SPARSITY

As we have shown, solving MAP inference queries subject to parity constraints is hard in general. However, adding parity constraints can sometimes makes the optimization easier. For example, when $A$ is the identity matrix, enforcing $A\sigma = b \mod 2$ corresponds to fixing the values of all variables and leads to a trivial optimization problem. Empirically, **sparse** constraints, such as the ones used in low density parity check (LDPC) codes from Gallager [8], tend to be much easier to solve. Unfortunately, constructions in both Propositions 1 and 2 to create pairwise independent hash functions require parity constraints that are of average length $n/2$, i.e., the corresponding matrix $A$ is not sparse.

A set of parity constraints specified through matrices $A, b$ defines a set of solutions $\mathcal{S} = \{x \in \{0, 1\}^n : Ax = b\}$, which is the translated nullspace of the matrix $A$. The nullspace is a vector space, defined with operations over the finite field $\mathbb{F}(2)$, i.e. modular arithmetic. Exploiting basic linear algebraic properties, it can be shown that applying **elementary row operations** to $[A|b]$ does not change the solution set $\mathcal{S}$ and thus the optimization problem. On the other hand, the parity polytope we described earlier is *not* a function of the solution set $\mathcal{S}$ but *depends explicitly* on the form of the matrices $A$ and $b$. This fact was also noted by Feldman et al. [7], who showed that a new matrix $[A'|b']$ constructed from $[A|b]$ by adding new rows that are linear combinations of the rows of $[A|b]$ can lead to a tighter LP relaxation, although $Ax = b$ and $A'x = b'$ define the same solution set $\mathcal{S}$ (because the constraints added are all implied).

In this paper we propose to exploit these facts and rewrite the constraints in a form that is equivalent, i.e., defines the same set of solutions, but is easier to solve. Specifically, given a a set of parity constraints specified through matrices $A, b$ we look for matrices $A', b'$ that define the same set of solutions, namely $\{x \in \{0, 1\}^n : Ax = b\} = \{x \in \{0, 1\}^n : A'x = b'\}$ but are much sparser, namely $||[A'|b']||_1 \ll ||[A|b]||_1$. Unfortunately, even finding a sparse linear combination of the rows is computationally intractable, as it can be seen as an instance of MAXIMUM-LIKELIHOOD DECODING, where the code is given in terms of the generators (the rows of $A$) rather than the check matrix. We therefore propose to use two approaches:

- Perform Gauss-Jordan elimination on $[A|b]$ to convert $[A|b]$ to **reduced row echelon form**;

- Try all combinations of up to $k$ rows $r_1, \cdots, r_k$ of $[A|b]$, and if their sum $r_1 \oplus \cdots \oplus r_k$ is sparser than any of the $r_i$, substitute $r_i$ with $r_1 \oplus \cdots \oplus r_k$.

Both techniques are based on **elementary row operations** and therefore are guaranteed to maintain the solution set $\mathcal{S}$ and to improve sparsity.

In Figure 2 we show the median upper and lower bounds found by CPLEX for several randomly generated constraints on a random $10 \times 10$ Ising grid model with mixed interactions. Starting with a matrix $A$ generated using the Toeplitz matrix construction in Proposition 2, we run CPLEX for 10 minutes with and without sparsification, reporting the best upper and lower bounds found. We see that without any preprocessing (NoPre) CPLEX fails at finding any integer solution when there are more than 15 parity constraints. Performing Gauss-Jordan elimination (Diag) significantly improves both the upper bound and the lower bound. The effect is particularly significant for a large number of constraints, when the reduced row echelon form of $A$ is close to the identity matrix. Adding the additional greedy substitution step (DiagGreedy, looking at all combinations of up to $k = 4$ rows) slightly improves the quality of the upper bound, but the lower bound significantly degrades. Therefore, for the rest of the paper we will use only Gauss-Jordan elimination preprocessing.

## 6 LOWER BOUNDS: SHORT XORS

As mentioned in Section 3.2, one practical way to obtain lower bounds from the WISH algorithm is to use suboptimal solutions of the underlying MAP inference problems. Here we explore a different way, namely, using sparse or *short* parity constraints (XORs), which are often easier for constraint solvers to reason about. This results in a family of hash functions that is uniform *but not pairwise independent*, leading to a weaker but practically valuable version of Theorem 1.

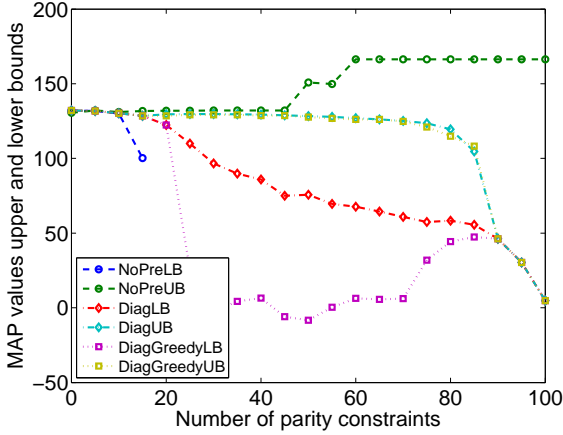

Figure 2: Upper and lower bounds with and without sparsification.

### 6.1 WISH WITH UNIFORM HASHING

Fix *any* subset $\mathcal{A}^{m \times n} \subseteq \{0,1\}^{m \times n}$ of $m \times n$ matrices. We will later create short XORs by choosing $\mathcal{A}^{m \times n}$ such that every row of every matrix in this set has only $k \ll n/2$ non-zero entries.

**Proposition 3.** *Let $A \in \mathcal{A}^{m \times n}$ and $b \in \{0,1\}^m$. The family $\mathcal{H}^{n,m} = \{h_{A,b}(x) : \{0,1\}^n \to \{0,1\}^m\}$ where $h_{A,b}(x) = Ax + b \mod 2$ is a family of **uniform** hash functions.*

*Proof.* Let $x \in \{0,1\}^n$, $A \in_R \mathcal{A}^{m \times n}$, $b \in_R \{0,1\}^m$, and $a_i$ denote the $i$-th row of $A$. Then, for all $i$:

$$\Pr[a_i x \oplus b_i = 0] = \sum_{v \in \{0,1\}^n} \Pr[a_i = v] \Pr[vx \oplus b_i = 0]$$

$$= \sum_{v \in \{0,1\}^n} \Pr[a_i = v] \Pr[b_i = vx \mod 2]$$

$$= \sum_{v \in \{0,1\}^n} \Pr[a_i = v] \frac{1}{2} = \frac{1}{2}$$

Hence, $\Pr[Ax + b = 0 \mod 2] = \prod_i \Pr[a_i x + b_i = 0 \mod 2] = \frac{1}{2^m}$ for any $x$, proving uniformity. □

With such a family of hash functions, Theorem 1 as such does not hold, but the following weaker, one-directional version still does:

**Theorem 3.** *For any $\delta > 0$, positive constant $\alpha \leq 0.0042$, and families $\mathcal{H}^{n,i}$ of uniform (but not necessarily pairwise independent) hash functions, with probability at least $(1-\delta)$, WISH($\{\mathcal{H}^{n,i}\}$) outputs an estimate no larger than $16Z = 16 \sum_{\sigma \in \mathcal{X}} w(\sigma)$.*

In other words, even without pairwise independence, the output divided by 16 is a lower bound with high probability. To prove this result, we employ a proof strategy similar to the one used by Ermon et al. [6].

For completeness, we start by stating some definitions we borrow from that work:

**Definition 4** ([6]). Fix an ordering $\sigma_i, 1 \leq i \leq 2^n$, of the configurations in $\mathcal{X}$ such that for $1 \leq j < 2^n$, $w(\sigma_j) \geq w(\sigma_{j+1})$. For $i \in \{0, 1, \cdots, n\}$, define $b_i \triangleq w(\sigma_{2^i})$. Define a special *bin* $B \triangleq \{\sigma_1\}$ and, for $i \in \{0, 1, \cdots, n-1\}$, define *bin* $B_i \triangleq \{\sigma_{2^i+1}, \sigma_{2^i+2}, \cdots, \sigma_{2^{i+1}}\}$.

Next we prove a new bound on $M_i$ that holds **regardless of pairwise independence**:

**Lemma 1.** *Suppose $h_{A,b}^i$ is chosen from a family $\mathcal{H}^{n,i}$ of universal (but not necessarily pairwise independent) hash functions. Let $M_i = \text{Median}(w_i^1, \cdots, w_i^T)$ be defined as in Algorithm 1 and $b_i$ as in Definition 4. Then, for all $c \geq 2$, there exists an $\alpha^*(c) > 0$ such that for $0 < \alpha \leq \alpha^*(c)$,*

$$\Pr\left[M_i \leq b_{\max\{i-c, 0\}}\right] \geq 1 - \exp(-\alpha T)$$

*Proof.* The statement trivially holds when $i - c \leq 0$. Otherwise, let us define the set of the $2^j$ heaviest configurations as in Definition 4, $\mathcal{X}_j = \{\sigma_1, \sigma_2, \cdots, \sigma_{2^j}\}$. Define the following random variable $S_j(h_{A,b}^i) \triangleq \sum_{\sigma \in \mathcal{X}_j} 1_{\{A\sigma = b \mod 2\}}$ which gives the number of elements of $\mathcal{X}_j$ satisfying the random parity constraints $A\sigma = b \mod 2$. The randomness is over the choice of $A$ and $b$ when $h_{A,b}^i$ is sampled from $\mathcal{H}^{n,i}$. Since $\mathcal{H}^{n,i}$ is a family of uniform hash functions, by definition for any $\sigma$ the random variable $1_{\{A\sigma = b \mod 2\}}$ is Bernoulli distributed with probability $1/2^i$. Then it follows that $\mathbb{E}[S_j(h_{A,b}^i)] = \sum_{\sigma \in \mathcal{X}_j} 1/2^i = \frac{|\mathcal{X}_j|}{2^i} = 2^{j-i}$.

The random variable $w_i$ is defined as $w_i = \max_\sigma w(\sigma)$ subject to $A\sigma = b \mod 2$. Then we have:

$$\Pr[w_i \leq b_j] = \Pr[w_i \leq w(\sigma_{2^j})] \geq \Pr[S_j(h_{A,b}^i) < 1]$$

which is the probability that no configuration from $\mathcal{X}_j$ satisfies $i$ randomly chosen parity constraints. Notice that $S_j(h_{A,b}^i)$ is non-negative, hence from Markov's Inequality, $\Pr[S_j(h_{A,b}^i) \geq 1] \leq \mathbb{E}[S_j(h_{A,b}^i)] = 2^{j-i}$. Thus for $j = i - c$ and $c \geq 2$, we have:

$$\Pr[w_i \leq b_{i-c}] \geq \Pr[S_{i-c}(h_{A,b}^i) < 1] \geq 1 - 2^{-c} \geq 3/4$$

Finally, since $w_i^1, \cdots, w_i^T$ are i.i.d. realizations of $w_i$, we can apply Chernoff's Inequality to the corresponding indicator variables $I_t = I(w_i^t \leq b_{i-c})$ each with mean $\geq 3/4$ and obtain:

$$\Pr[M_i \leq b_{i-c}] = \Pr\left[\sum_t I_t \geq T/2\right] \geq 1 - \exp(-\alpha^*(c)T)$$

where $\alpha^*(2) = 2(3/4 - 1/2)^2 = 1/8$. □

With this new lemma, we have all we need to prove Theorem 3. The proof is similar to the one of Theorem 1 and is not included for space reasons.

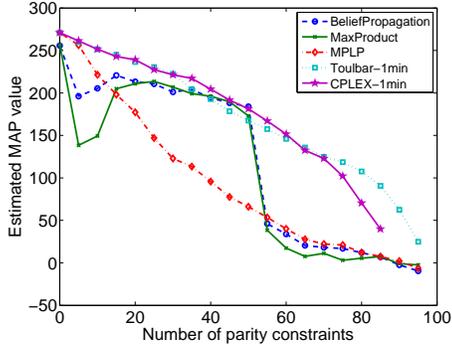

(a) Attractive $10 \times 10$. Length 4 Xors

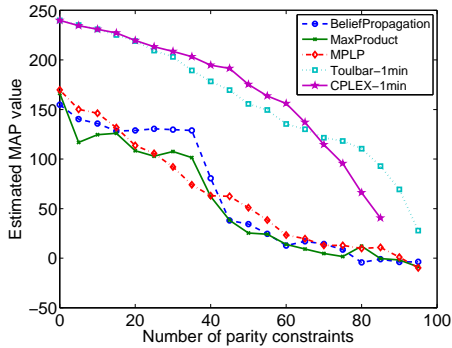

(b) Mixed $10 \times 10$. Length 4 Xors

Figure 3: Optimization with short parity constraints.

## 6.2 EVALUATION OF SHORT XORS

As briefly alluded to earlier, we will use *short XORs* in order to make MAP inference more efficient in practice. Specifically, we will choose $A$ uniformly from $\mathcal{A}_k^{m \times n} \subseteq \{0, 1\}^{m \times n}$, which is the set of matrices such that every row has only $k$ non-zero entries, where $k$ will typically be much smaller than $n/2$. In general, smaller values of $k$ lead to faster execution of WISH but at the cost of weaker lower bounds.

Figure 3 compares several approaches to solve the MAP inference problems constrained by random parity constraints for a $10 \times 10$ Ising grid model with attractive and mixed interactions (external field $f = 1.0$ and weight $w = 3.0$; see below for a formal description of the probabilistic model used). We compare three message passing approaches, namely Belief Propagation (BP), Max-Product (MP), and MPLP [9], and two combinatorial optimization solvers, namely Toulbar [1] and CPLEX 12.3 [15], both with a 1 minute time limit. We show the median value of the solution found over 50 realizations, for each number of parity constraints added. We run the Message Passing methods until they find a feasible solution satisfying the parity constraints or up to 10000 iterations. If no feasible solution is found, we round the final beliefs to an integer solution and project it on the feasible set by solving the linear equations with Gaussian Elimination, thus changing the value of some of the variables. For this problem, using "long" parity constraints of length 50, Message Passing methods can only find feasible solutions for up to 10 constraints (consistent with CPLEX performance in Figure 2). In contrast, as shown in Figure 3, using short XORs of length 4 (typical values encountered e.g. for low density parity check codes), Message Passing methods can find feasible solutions for up to about $40-50$ constraints, at which point there is a significant performance drop caused by the need for a projection step. We see that for the attractive case, Message Passing methods are competitive with combinatorial optimization approaches but only for a moderate number of constraints. In the more challenging mixed interactions case, CPLEX and Toulbar appear to be clearly superior. We think the the unsatisfactory performance of message passing techniques (compared e.g. to when used for LDPC decoding) is caused by the more complicated probabilistic dependencies imposed by the Ising model, which is much more intricate than a typical transmission error model.

## 7 EXPERIMENTS

We evaluate the performance of WISH augmented with Toeplitz-matrix based hash functions (from Proposition 2) and CPLEX 12.3 [15] to solve the ILP formulation of the MAP queries. All the optimization instances are solved in parallel on a compute cluster, with a timeout of 10 minutes on Intel Xeon 5670 3GHz machines. We use Gauss-Jordan elimination preprocessing to improve the quality of the LP relaxations. We use the Jaroslow encoding for parity constraints $j \in \mathcal{J}$ such that $|N(j)| \leq 10$, and the Yannakakis encoding otherwise. We evaluate the lower bound and upper bounds for the partition functions of $M \times M$ grid Ising models for $M \in \{10, 15\}$, with random interactions (positive and negative) and external field $f \in \{0.1, 1.0\}$. Specifically, there are $M^2$ binary variables, with single node potentials $\psi_i(x_i) = \exp(f_i x_i)$ and pairwise interactions $\psi_{ij}(x_i, x_j) = \exp(w_{ij} x_i x_j)$, where $w_{ij} \in_R [-w, w]$ and $f_i \in_R [-f, f]$.

We compare with Loopy BP [23] which estimates $Z$, Tree Reweighted BP [30] which gives a provable upper bound, and the Mean Field approach [31] which gives a provable lower bound. We use the implementations in the LibDAI library [22] and compare with ground truth obtained using the Junction Tree method [20].

Figure 4 shows the error in the resulting estimates, together with the upper and lower bounds obtained with WISH augmented with Toeplitz-matrix hashing and CPLEX. We immediately see that our *lower bounds*

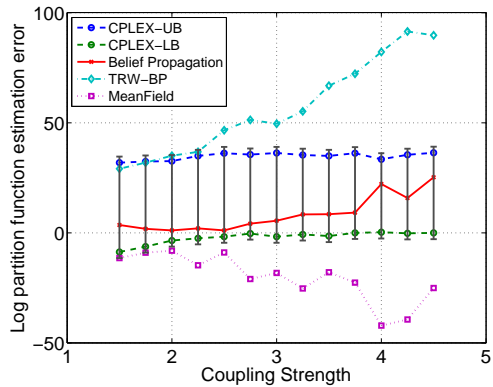
(a) Mixed $10 \times 10$. Field 0.1.

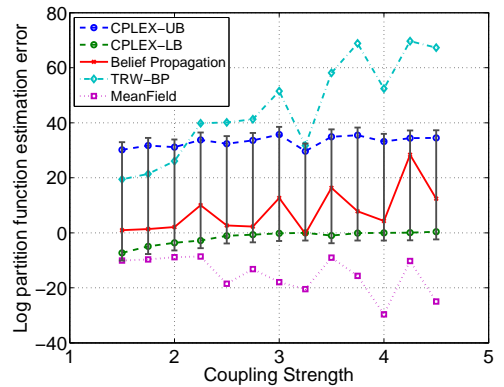
(b) Mixed $10 \times 10$. Field 1.0.

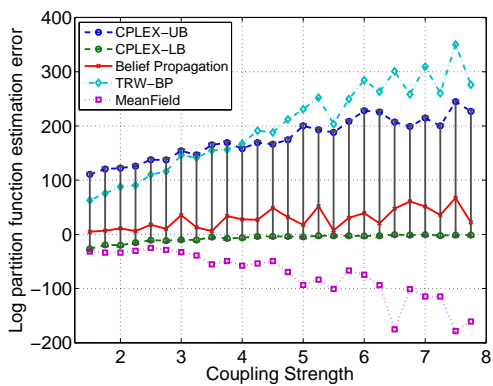
(c) Mixed $15 \times 15$. Field 0.1.

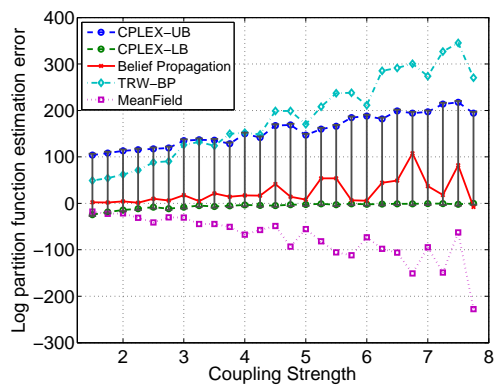
(d) Mixed $15 \times 15$. Field 1.0.

Figure 4: Results on spin glasses grids.

are highly accurate (error close to 0), which means that the lower bounds provided by CPLEX for the ILPs must be close to optimality. Similarly good lower bounds can also be obtained using the original WISH algorithm [6], and also using SampleSearch [11]. However, neither SampleSearch nor the original WISH (without LP relaxations) provide *upper bound* guarantees, only the TRWBP approach does. Specifically, the original WISH algorithm with Toulbar [1] provides an upper bound only upon proving optimality for all optimization instances in the inner loop. In contrast, the ILP formulation provides us with **anytime** and gradually improving upper bounds based on LP relaxations (cf. Figure 1), often well before it can actually solve the problems to optimality (which might not be possible on larger instances) or, in principle, even before it can find a feasible solution. Figure 4 shows that our upper bounds are significantly tighter than the ones obtained using TRWBP in the hard weights region. Further, our ILP approach is guaranteed to eventually give an accurate answer, within a constant factor, given enough time. In contrast, message passing techniques are usually quite fast (if they converge) but do not provide better results with more runtime.

## 8 CONCLUSIONS

We explored several extensions of the recent WISH [6] algorithm for computing discrete integrals. First, we used a better, more deterministic and thus more stable construction for pairwise independent hash functions. Using a connection with max-likelihood decoding of binary codes, we showed that the MAP inference queries generated by WISH are in general not polynomial time solvable or even approximable. On the positive side, this led to the use of an ILP formulation for the problem, inspired by iterative message passing decoding. To increase the practicality of the ILP approach, we sparsified parity constraints while preserving their desirable properties. Further, we extended WISH to directly utilize uniform but not necessarily pairwise independent hash functions, leading to computationally easier optimization problems while still providing probabilistic lower bound guarantees. Finally, we showed that by solving a sequence of LP relaxations we can obtain not only very accurate lower bounds but also upper bounds that are much tighter than the ones provided by TRWBP, which is based on tree decomposition and convexity.